%% file: main.tex
\PassOptionsToPackage{x11names,table}{xcolor}
\documentclass[10pt,twocolumn,letterpaper]{article}

 \usepackage{iccv}              
\usepackage[accsupp]{axessibility}
\input{preamble}

\definecolor{iccvblue}{rgb}{0.21,0.49,0.74}
\usepackage[pagebackref,breaklinks,colorlinks,allcolors=iccvblue]{hyperref}
\usepackage{soul}
\def\paperID{4513} 
\def\confName{ICCV}
\def\confYear{2025}

\input{macro}

\title{\Ours{}: Compact and Fast 3D Feature Fields}

\author{
  Hyunjoon Lee\quad\quad
  Joonkyu Min\quad\quad
  Jaesik Park\footnotemark[1]\\
  Seoul National University, Republic of Korea\\
  {\tt\small \{hjlee4772, timothy0609, jaesik.park\}@snu.ac.kr}
}

\begin{document}
\twocolumn[{%
    \renewcommand\twocolumn[1][]{#1}%
    \maketitle
    \centering
    \vspace{-3mm}
    \newcommand{\teaserwidth}{0.8\textwidth}
    \includegraphics[width=0.95\textwidth]{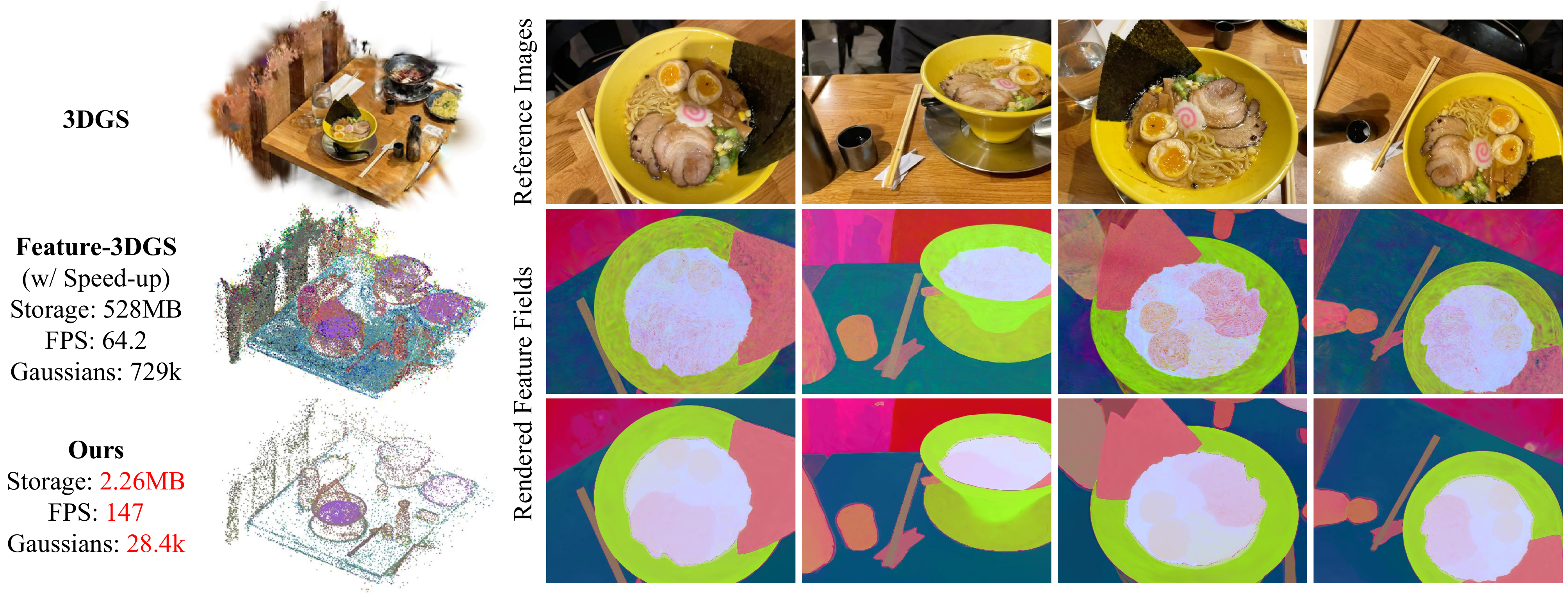}
    \captionsetup{type=figure,hypcap=false}
    \vspace{-4mm}
    \captionof{figure}{We propose \Ours{} for constructing a compact and fast 3D feature field from 3D Gaussians.
    The previous method (Feature-3DGS) jointly optimizes features with colors, resulting in excessive Gaussians for rendering the feature field. \Ours{} effectively compresses and sparsifies the 3D feature field while maintaining sufficient details as shown in the rendered feature maps. 
    \label{fig:teaser}
    }%
    \vspace{1.5em}
}]
\renewcommand{\thefootnote}{\fnsymbol{footnote}}
\footnotetext[1]{Corresponding author.}
\input{sec/0_abstract}    
\input{sec/1_intro}
\input{sec/2_related}
\input{sec/3_method}
\input{sec/4_Experiments}
\input{sec/5_conclusion}
\section*{Acknowledgements. }
This work was supported by IITP grant (RS-2021-II211343: AI Graduate School Program at Seoul National Univ. (5\%) and RS-2023-00227993: Detailed 3D reconstruction for urban areas from unstructured images (60\%)) and NRF grant (No.2023R1A1C200781211 (35\%)) funded by the Korea government (MSIT). 

{
    \small
    \bibliographystyle{ieeenat_fullname}
    \bibliography{main}
}

\clearpage
\input{sec/X_supple}

\end{document}

%% file: preamble.tex
\usepackage{multirow}
\usepackage{algorithm}
\usepackage[noend]{algpseudocode}
\algrenewcommand{\algorithmicindent}{1.0em}
\usepackage[hang,flushmargin,symbol]{footmisc}
\usepackage{graphicx}	
\usepackage{amsmath}	
\usepackage{amssymb}	
\usepackage{booktabs}
\usepackage{times}
\usepackage{microtype}
\usepackage{epsfig}
\usepackage{caption}
\usepackage{float}
\usepackage{placeins}
\usepackage{color, colortbl}
\usepackage{stfloats}
\usepackage{enumitem}
\usepackage{tabularx}
\usepackage{xstring}
\usepackage{xspace}
\usepackage{url}
\usepackage{subcaption}
\usepackage[]{xcolor}
\usepackage{pifont}
\usepackage{cuted}  
\usepackage{caption}

%% file: macro.tex
\definecolor{softblue}{HTML}{1f77b4}
\definecolor{softgreen}{HTML}{2ca02c}
\definecolor{softred}{HTML}{d62728}
\definecolor{softpurple}{HTML}{9467bd}

\newcommand{\Ours}{$\text{CF}^3$}

\definecolor{colorf}{HTML}{FFC1C3}
\definecolor{colors}{HTML}{FFC991}
\definecolor{colort}{HTML}{FFF6A9}

\newcommand{\boxcolorf}[1]{%
  \leavevmode%
  \begingroup\setlength{\fboxsep}{1pt}%
  \colorbox{colorf}{\hspace*{2pt}\vphantom{Ay}#1\hspace*{2pt}}%
  \endgroup
}

\newcommand{\boxcolors}[1]{%
  \leavevmode%
  \begingroup\setlength{\fboxsep}{1pt}%
  \colorbox{colors}{\hspace*{2pt}\vphantom{Ay}#1\hspace*{2pt}}%
  \endgroup
}

\newcommand{\boxcolort}[1]{%
  \leavevmode%
  \begingroup\setlength{\fboxsep}{1pt}%
  \colorbox{colort}{\hspace*{2pt}\vphantom{Ay}#1\hspace*{2pt}}%
  \endgroup
}

%% file: sec/0_abstract.tex
\begin{abstract}
3D Gaussian Splatting (3DGS) has begun incorporating rich information from 2D foundation models. However, most approaches rely on a bottom-up optimization process that treats raw 2D features as ground truth, incurring increased computational costs. 
We propose a top-down pipeline for constructing compact and fast 3D Gaussian feature fields, namely, \Ours{}. We first perform a fast weighted fusion of multi-view 2D features with pre-trained Gaussians.
This approach enables training a per-Gaussian autoencoder directly on the lifted features, instead of training autoencoders in the 2D domain.
As a result, the autoencoder better aligns with the feature distribution.
More importantly, we introduce an adaptive sparsification method that optimizes the Gaussian attributes of the feature field while pruning and merging the redundant Gaussians, constructing an efficient representation with preserved geometric details. 
Our approach achieves a competitive 3D feature field using as little as 5\% of the Gaussians compared to Feature-3DGS.
\end{abstract}

%% file: sec/1_intro.tex
\section{Introduction}
\label{sec:intro}
Recent advances in 3D scene reconstruction have achieved significant progress in rendering high-fidelity images and precise 3D models, as exemplified by methods such as NeRF~\cite{mildenhall2021nerf} and 3DGS~\cite{kerr2023lerf}. 
With these advances, modern methods have aimed to integrate rich information from 2D foundation models, like CLIP~\cite{radford2021learning}, LSeg~\cite{li2022language}, and SAM~\cite{kirillov2023segment} into 3D representations. These methods extract patch-level or pixel-level features from multi-view images, including those designed for semantic understanding. In the case of semantic features, the extracted representations are distilled into the 3D space, forming a language or semantic 3D field capable of handling open-vocabulary queries, \eg, \lq wall\rq, \lq sofa\rq, \lq chair\rq, in real-time.

Prior works in this category~\cite{zhou2024feature,qin2024langsplat} typically optimize the embedding of semantic features, akin to learning color via photometric loss, across all Gaussians using multi-view raw visual feature maps. 
Since this joint color and feature learning strategy forces the recovery of color details with an excessive number of Gaussians, the resulting feature fields are often heavy and redundant.
Furthermore, directly embedding high-dimensional language features into 3D Gaussians incurs significant storage and computational costs. Several methods have been proposed to address these issues. 
For example, feature compression using autoencoders~\cite{qin2024langsplat} or decoder-only reconstruction~\cite{zhou2024feature}, as well as hash-grid techniques~\cite{zuo2024fmgs} and vector quantization~\cite{shi2024language}, have been explored.
However, these methods~\cite{zhou2024feature,qin2024langsplat,zuo2024fmgs,shi2024language} do not explicitly consider that Gaussians optimized for color may be redundant for expressing a feature field. 
In addition, previous feature embedding methods ~\cite{zhou2024feature,zuo2024fmgs,qin2024langsplat,shi2024language,wu2024opengaussian,liang2024supergseg} rely on raw features from 2D foundation models, which often lack multi-view consistency~\cite{el2024probing, chen2025feat2gs}. 

\input{figs/overview}

We propose an approach to eliminate redundant Gaussians and achieve high-quality feature fields. An overview of the compactness of our method is shown in~\cref{fig:teaser}.
\Cref{fig:pipeline} provides an overview of our pipeline, illustrating the stages of feature lifting, compression, and adaptive sparsification.

Similar to 3D-aware training in FiT3D~\cite{yue2024improving} and CONDENSE~\cite{zhang2024condense}, we first compute a weighted combination of 2D features in 3D, namely feature lifting, with a pre-trained 3DGS.
This scheme quickly achieves feature quality comparable to results from approaches that jointly optimize images and features.
We employ these spatially coherent and view-consistent rendered features as reference features. 

Moreover, unlike Feature-3DGS~\cite{zhou2024feature} and LangSplat~\cite{qin2024langsplat} that learn a per-pixel decoder, we suggest lifting the feature first (to get reference features) and then compressing it using a per-Gaussian autoencoder.
Since each Gaussian is directly assigned a fused and view-consistent reference feature, our method avoids the need for pre-compression of 2D feature maps, enabling direct training of the autoencoder for each Gaussian. Combined with variance filtering, this approach effectively removes inaccurate features that may arise during the lifting process, ensuring more reliable feature extraction.

Building on this compression, we propose an adaptive sparsification process to optimize the Gaussian feature field even further. This step optimizes Gaussian attributes and merges redundant Gaussians in stable regions. 
Here, stable regions refer to areas with a small gradient that already represent the scene well, making further refinement unnecessary. 

\noindent We summarize our main contributions below:
\begin{itemize}
    \item 
    We build a compact 3D feature field by lifting features via a pre-trained 3DGS and compressing them with a per-Gaussian autoencoder. This ensures robustness across downstream tasks since each Gaussian directly encodes view-consistent reference features.
    \item Our adaptive sparsification step optimizes the Gaussian feature field even further, which involves pruning and merging redundant Gaussians, while preserving essential details. As a result, our method achieves competitive performance while using as little as 5\% of the original number of Gaussians, improving storage efficiency and rendering speed.
\end{itemize}

%% file: figs/overview.tex
\begin{figure*}[t!]
\centering
\includegraphics[width=0.9\textwidth]{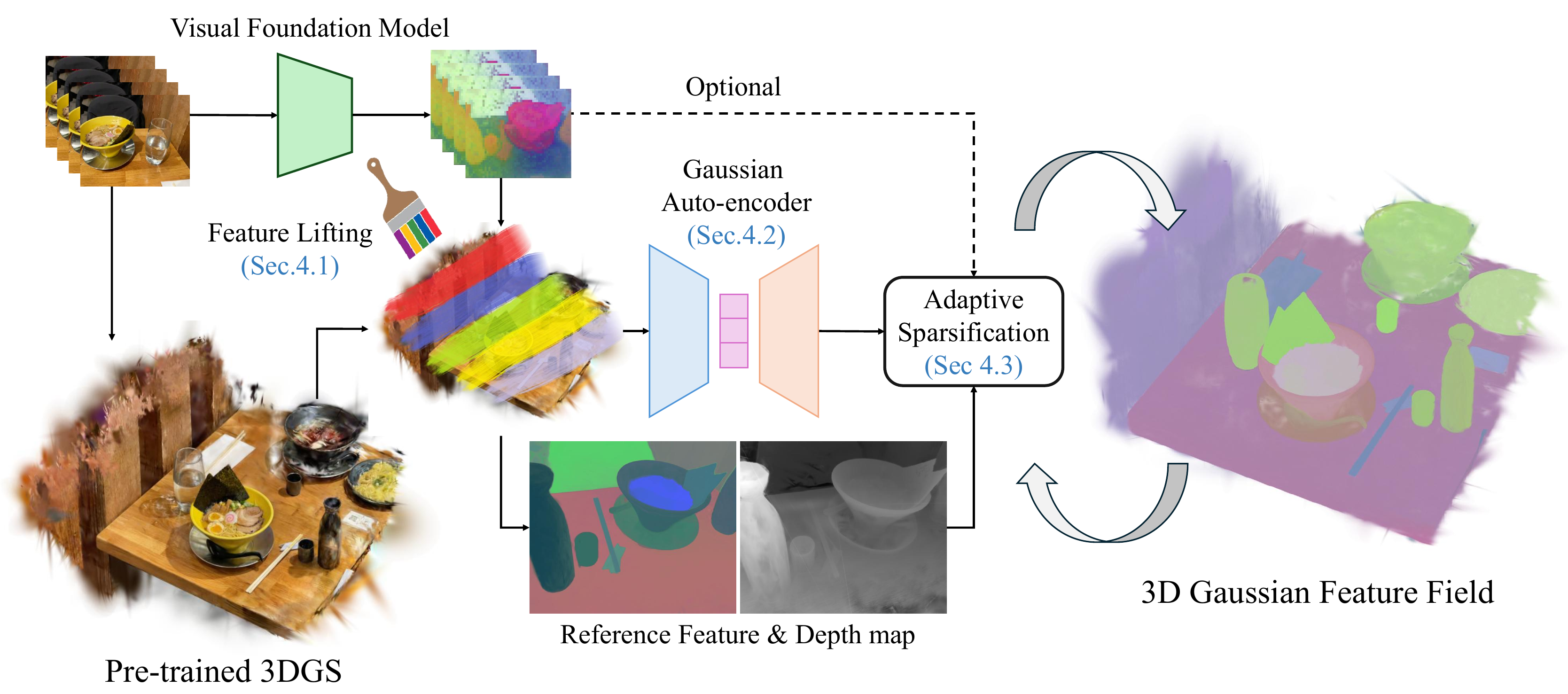}
    \caption{\textbf{Overview of our \Ours{} pipeline.} We utilize pre-trained 3D Gaussians to construct a 3D feature field. We adopt a weighted-sum strategy to lift features extracted from a visual foundation model into 3D. 
    Subsequently, a per-Gaussian autoencoder compresses high-dimensional features into lower-dimensional embeddings, effectively removing noisy features through a variance filtering step. 
    Afterward, adaptive sparsification merges redundant Gaussians, efficiently reducing the total Gaussian count and resulting in a compact 3D feature field.}
    \label{fig:pipeline}
\end{figure*}

%% file: sec/2_related.tex
\section{Related Work}
\label{sec:related}

\subsection{Visual feature embedding with NeRF}
Neural Radiance Fields (NeRF)-based approach pioneered beyond basic scene reconstruction by incorporating high-dimensional features extracted from 2D vision foundation models into 3D representations. By embedding features in NeRF, tasks such as semantic segmentation, object decomposition, language-based querying, and editing are enabled. 

These feature-embedded approaches can be broadly categorized into three groups. 
First, some approaches distill large-scale 2D embeddings (e.g., CLIP, DINO) into 3D fields for open-vocabulary queries or text-driven object segmentation ~\cite{liu2023weakly,kobayashi2022decomposing,kerr2023lerf,tschernezki2022neural,engelmann2023open}. 
They typically employ multi-scale patch extraction or pixel-aligned semantic features~\cite{li2022language,ghiasi2022scaling}, combined with feature alignment and additional losses (e.g., regularization) to enhance geometry and segmentation.
Second, several approaches introduce object-level decomposition or local NeRF blocks for scene editing or refining sparse/noisy 2D annotations~\cite{wang2022dm,zhi2021place,zhang2023nerflets}. They achieve higher interpretability and efficient object manipulation through targeted object fields, specialized losses, or local MLPs.
Finally, some methods address panoptic labeling in 3D by fusing bounding primitives or 2D panoptic masks with NeRF~\cite{siddiqui2023panoptic,fu2022panoptic,kundu2022panoptic}.

\subsection{Visual feature embedding with 3DGS}
3D Gaussian splatting~\cite{kerbl20233d} has demonstrated strong performance in real-time novel view rendering by representing scenes explicitly via anisotropic Gaussians, which can be rasterized and blended at high speed. 
To further enhance these representations with semantic information, several works~\cite{zhou2024feature,zuo2024fmgs,qin2024langsplat} have proposed integrating features from 2D foundation models. Early efforts employ optimization-based feature distillation, where embeddings (e.g., from CLIP, DINO, or SAM) are lifted into 3D space through iterative optimization.
Subsequent approaches~\cite{shi2024language,wu2024opengaussian,liang2024supergseg} address the storage overhead of large embeddings by quantizing or compressing features, or by clustering Gaussians into superpoint-like structures. 
A few methods adopt training-free schemes, aggregating 2D features into 3D with a weighted average method rather than explicit backpropagation~\cite{marrie2024ludvig,cheng2024occam}.
Others~\cite{wang2024gsemsplat,fan2025large} attempt a feedforward model that can process sparse or unposed images and generate a feature-embedded Gaussians in a single pass. 

\subsection{Reducing storage overhead}
Recent research on 3DGS focuses on reducing storage overhead while maintaining quality through three complementary strategies: (1) compressing individual Gaussian attributes through vector quantization or selective spherical harmonics (SH) pruning ~\cite{fan2023lightgaussian, niedermayr2024compressed, wang2024end}, (2) reorganizing scenes with structured encodings (anchor-based or hash-grid-based) to leverage spatial coherence ~\cite{chen2024hac, lee2024compact, Shin_Localityaware_ICLR_2025,lu2024scaffold}, and (3) adaptively controlling splat density by pruning less significant Gaussians or densifying under-reconstructed regions ~\cite{cheng2024gaussianpro, girish2024eagles, ren2024octree, wang2024end, zhang2024pixel}.
In attribute compression, highly correlated features, such as scale, rotation, or SH color coefficients, are typically clustered into codebooks to reduce redundancy, while low-bit quantization and optional re-encoding with standard codecs further reduce storage requirements~\cite{morgenstern2024compact, xie2024mesongs,Shin_Localityaware_ICLR_2025}. Structured representations organize splats in anchor-based clusters, 2D grids, employ octrees, or Morton ordering to efficiently skip empty regions~\cite{fang2024mini, kheradmand20243d, morgenstern2024compact, xie2024mesongs}, sometimes replacing SH with learned MLPs~\cite{lee2024compact}.
Pruning strategies eliminate overlapping or negligible splats~\cite{chen2024hac, fan2023lightgaussian, girish2024eagles, lee2024compact, wang2024end}, while selective densification enhances fine details using multi-view gradients, as well as depth or normal cues ~\cite{cheng2024gaussianpro, ren2024octree, zhang2024pixel}. 

%% file: sec/3_method.tex
\section{Preliminary}
3DGS scene $\mathcal{S}=\{g_i | i=1, \cdots, N\}$ is represented with $N$ Gaussians, where each Gaussian has a center coordinate $\boldsymbol{\mu}\in \mathbb{R}^3$, a covariance matrix ${\Sigma}\in \mathbb{R}^{3\times 3}$, an opacity $\alpha\in\mathbb{R}_+$. \begin{equation}
    g_i(\mathbf{x}) = \exp\left(-\frac{1}{2}(\mathbf{x}-\boldsymbol{\mu})^\top\boldsymbol{\Sigma}^{-1}(\mathbf{x}-\boldsymbol{\mu})\right)
.
\end{equation}
Color $\boldsymbol{c}$ at each pixel in the image is rendered via alpha blending of Gaussian's color or spherical harmonics features $\boldsymbol{c}_i$ considering depth order to the viewpoint. Similarly, depth is rendered by weighting each Gaussian with distance 
$d_i$, defined as the distance from the camera center to each Gaussian~\cite{kerbl20233d}.
\begin{equation}
    {C}  = \sum_{i=1}^N \boldsymbol{c}_i\alpha_i \prod^{i-1}_{j=1}(1-\alpha_j) =\sum_{i=1}^N \boldsymbol{c}_i \alpha_i T_i =\sum_{i=1}^N \boldsymbol{c}_iw_i,
\label{eq:3dgs_color}
\end{equation}
\begin{equation}
    D  = \sum_{i=1}^N d_i\alpha_i \prod^{i-1}_{j=1}(1-\alpha_j) =\sum_{i=1}^N d_i \alpha_i T_i =\sum_{i=1}^N d_iw_i,
\label{eq:3dgs_depth}
\end{equation}
where $T_i\in\mathbb{R}_+$ is transmittance. We denote $w_i$ as the \emph{weight of the corresponding Gaussian} contributing to each pixel.

\section{Method}
\label{sec:method}
\subsection{Feature Lifting}\label{sec:featurelifting}
The prior methods optimize the features during 3DGS~\cite{zhou2024feature,zuo2024fmgs} training, which result in a long training time, making it difficult to scale up.
We use an alternative and scalable solution to lift visual features to our 3DGS scene.
Given $M$ images, $P$ pixels each, let's assume we have image features $\boldsymbol{F}_{m,p}$ for $p$-th pixel in $m$-th image, where $\|\boldsymbol{F}_{m,p}\|=1$. 
Let $\mathcal{G}_{m,p}$ be an index set of Gaussians that are projected onto pixel $p$ of image $m$.

The problem is to minimize the gap between the image features $\boldsymbol{F}_{m,p}$ and the rendered features $\sum_{i\in \mathcal{G}_{m,p}}w_{i,m,p}\boldsymbol{f}_{i}$, $\boldsymbol{f}_i$ indicates corresponding features for each 3D Gaussians with the constraint of $\|\boldsymbol{f}_i\|=1$. Here, $w_{i,m,p}$ refers to the weight introduced in \cref{eq:3dgs_color}.
The approximate solution is simply computing the weighted sum over a set of pixels that are included in a 2D splat of Gaussian $g_i$, noted as $\mathcal{P}_{i,m}$:
\begin{equation}
\begin{split}
\boldsymbol{f}_i \approx \frac{\sum_{m=1}^M \sum_{p\in \mathcal{P}_{i,m}} w_{i,m,p} \boldsymbol{F}_{m,p}}{\sum_{m=1}^M \sum_{p\in \mathcal{P}_{i,m}} w_{i,m,p}}.
\label{eq:approx_f}
\end{split}
\end{equation}
This idea appears in recent training-free feature aggregation methods~\cite{marrie2024ludvig, cheng2024occam}. 
As shown in~\cref{fig:main_multiview}, lifting visual features to 3D Gaussians can reduce multi-view inconsistencies~\cite{yue2024improving}.

\input{figs/main_multiview}

In addition, we can measure the variance of the approximated features. Without considering the covariance among feature dimensions, the variance of each $d$-th dimension of features can be computed as follows:
\begin{equation}
\begin{split}
    \text{Var}(\boldsymbol{f}_i)_d \approx \frac{\sum_{m=1}^M \sum_{p\in \mathcal{P}_{i,m}} \!w_{i,m,p} (\boldsymbol{F}_{m,p})_d^2}{\sum_{m=1}^M \sum_{p\in \mathcal{P}_{i,m}} \!w_{i,m,p}} \!-\! (\boldsymbol{f}_i)_d^2.
\end{split}
\label{eq:approx_var}
\end{equation}

Most of the Gaussians at accurate positions with consistent features have low variance. However, some 3D Gaussians with inaccurate geometry or those located at the edges of objects often average irrelevant information.
Therefore, we filter out $i$-th Gaussian whose norm of the approximated variance $\text{Var}(\boldsymbol{f}_i)$ is larger than the top 0.01\% for the downstream pipeline.

\subsection{Feature Compression}
Unlike the existing method~\cite{qin2024langsplat} that trains an autoencoder before feature lifting, we suggest lifting the features first and then compressing them using a per-Gaussian autoencoder. 

\input{figs/optimization}

As shown in ~\cref{fig:main_multiview}, 
our autoencoder is trained directly on the lifted features, making it better aligned with the actual feature distribution used during inference.
Note that our autoencoder (MLP with five layers: [128, 64, 32, 16, 3]-\textit{dims} for encoding) compresses the D-dimensional lifted features $\boldsymbol{f}$ into a just 3-dimensional latent space. Interestingly, this is equivalent to treating the encoded feature as 3-channel RGB colors.
This design allows us to leverage the existing 3DGS rasterizer directly, and the outputs can be directly decoded into semantic features.
Our autoencoder is trained with MSE loss, together with cosine-similarity loss and a lightweight similarity structure preserving regularizer.

\noindent The objective is defined as follows:
\begin{align}
    &\mathcal{L} = \mathcal{L}_{\mathit{MSE}} + \lambda_{\mathit{cos}} \cdot \mathcal{L}_{\mathit{cos}} + \lambda_{\mathit{struc}} \cdot \mathcal{L}_{\mathit{struc}}, \\
    &\mathcal{L}_{\mathit{MSE}} = \mathbb{E}_{i\in\mathcal{G}}\left[\left\| \mathcal{D}(\mathcal{E}(\boldsymbol{f}_i)) - \boldsymbol{f}_i \right\|_2\right ], \\
    &\mathcal{L}_{\mathit{cos}} = \mathbb{E}_{i\in\mathcal{G}}\left[1 - \frac{ \langle \mathcal{D}(\mathcal{E}(\boldsymbol{f}_i)),\ \boldsymbol{f}_i \rangle }{ \| \mathcal{D}(\mathcal{E}(\boldsymbol{f}_i)) \| \cdot \| \boldsymbol{f}_i \| }\right ], \\
    &\mathcal{L}_{\mathit{struc}} \!=\! \mathbb{E}_{i\neq j, (i,j)\in\mathcal{G}} \!\! \left[ \left\| \cos(\boldsymbol{f}_i, \boldsymbol{f}_j) \!-\! \cos(\mathcal{E}(\boldsymbol{f}_i),\mathcal{E}(\boldsymbol{f}_j)) \right\| \right ]\!,
\label{eq:ae_loss}
\end{align}
where $\mathcal{G}$ is a set of gaussians, $\mathcal{E}(\boldsymbol{f}_i)$ is the encoded latent feature, and $\mathcal{D}(\mathcal{E}(\boldsymbol{f}_i))$ is the corresponding reconstruction.

\subsection{Adaptive Sparsification}\label{sec:sparsification}
As a next step, we optimize the Gaussian attributes $(\boldsymbol{\mu}, \boldsymbol{\Sigma}, \alpha, \boldsymbol{f})$ in our Gaussian feature field, which involves iterative pruning and merging 3D Gaussians to reduce redundancy.
\Cref{fig:optimization} shows the overall sparsification pipeline.
The sparsification process uses the rendered reference feature $\boldsymbol{F}_{ref}$ and depth map $D_{ref}$ of the Gaussian feature field being optimized. They are obtained from \cref{eq:3dgs_color,eq:3dgs_depth}. Note that we can reuse \cref{eq:3dgs_color} for feature rendering since the compressed feature is 3-dimensional.
These view-consistent features stabilize optimization by providing supervision across views.
The depth regularization term encourages geometric consistency with the original scene structure, enabling better alignment between the pre-trained 3DGS and \Ours{}. 

We define the objective for optimizing our 3D Gaussian attributes as follows:
 \begin{align}
     &\mathcal{L} = \mathcal{L}_{f} + \lambda_{depth} \cdot \mathcal{L}_{depth},\\
     &\mathcal{L}_{f} = \|\boldsymbol{F}_{ref} - \boldsymbol{F} \|_1, \\
     & \mathcal{L}_{depth} = \|{D}_{ref} - D\|_1,
 \end{align}
where $\boldsymbol{F}$ is the rendered feature map followed by the trained decoder.
\input{tables/algo}

Our Gaussian field optimization involves the following adaptive sparsification steps.
(1) \emph{Pruning}. By following LightGaussian~\cite{fan2023lightgaussian}, we prune the 3D Gaussians based on the global contribution $C(g_i)$, which is the sum of the weights on each image pixel:
\begin{equation} \label{eq:contrib}
    C(g_i) = \sum_{m=1}^M\sum_{p\in \mathcal{P}_{i,m}}w_{i,m,p}.
\end{equation}

(2) \emph{Merging}. We then iteratively merge the neighboring pairs of Gaussians with the same semantic information. For each Gaussian, we identify its $k$-nearest neighbors and then choose pairs with a significant overlap. We measure the overlap between neighboring $i$-th and $j$-th 3D Gaussians using Mahalanobis distance $d_M$, 
\begin{equation}
d_M = (\boldsymbol{\mu}_i - \boldsymbol{\mu}_j)^\top \boldsymbol{\Sigma}^{-1}(\boldsymbol{\mu}_i - \boldsymbol{\mu}_j) < \chi^2_{\beta},
\label{eq:merge}
\end{equation}
which effectively quantifies the separation of two Gaussian distributions relative to their covariance, and uses it for deciding the pairs to be merged.

Inspired by moment matching method for Gaussian mixture reduction~\cite{schieferdecker2009gaussian}, the new attributes for 3D feature Gaussians  $(\boldsymbol{\mu}_{new}, \boldsymbol{\Sigma}_{new}, \alpha_{new}, \boldsymbol{f}'_{new})$ that approximately represents the two overlapping Gaussians $(\boldsymbol{\mu}_i, \boldsymbol{\Sigma}_i, \alpha_i, \boldsymbol{f}'_i)$, $(\boldsymbol{\mu}_j, \boldsymbol{\Sigma}_j, \alpha_j, \boldsymbol{f}'_j)$ are computed by the following equation:
\begin{align}
    \boldsymbol{\mu}_{new} &= \frac{\alpha_{i} \boldsymbol{\mu}_{i} + \alpha_{j} \boldsymbol{\mu}_{j}}{\alpha_{i} + \alpha_{j}},\\
    \boldsymbol{\Sigma}_{new} &= \frac{\alpha_{i}\!\left(\boldsymbol{\Sigma}_{i} \!+\! \boldsymbol{\mu}_{i} \boldsymbol{\mu}_{i}^\top\right) \!+\! \alpha_{j}\!\left(\boldsymbol{\Sigma}_{j} \!+\! \boldsymbol{\mu}_{j} \boldsymbol{\mu}_{j}^\top\right)}{\alpha_{i} \!+\! \alpha_{j}} - \boldsymbol{\mu}_{new} \boldsymbol{\mu}_{new}^\top,\\
    \alpha_{new} &= \alpha_{i} + \alpha_{j} - \alpha_{i} \alpha_{j},\\
    \boldsymbol{f}'_{new} &= \frac{\alpha_{i} \boldsymbol{f}'_{i} + \alpha_{j} \boldsymbol{f}'_{j}}{\alpha_{i} + \alpha_{j}},
\label{eq:merge}
\end{align}
where $\boldsymbol{f}' = \mathcal{E}(\boldsymbol{f}) \in \mathbb{R}^3$ denotes the latent feature compressed by the autoencoder $\mathcal{E}$.
Through our adaptive sparsification step, we construct a compact 3D feature field with significantly fewer Gaussians than the original 3DGS. The algorithm is summarized in~\cref{alg:sparsification}.

%% file: figs/main_multiview.tex
\begin{figure}[t!]
\centering
\includegraphics[width=0.9\columnwidth]{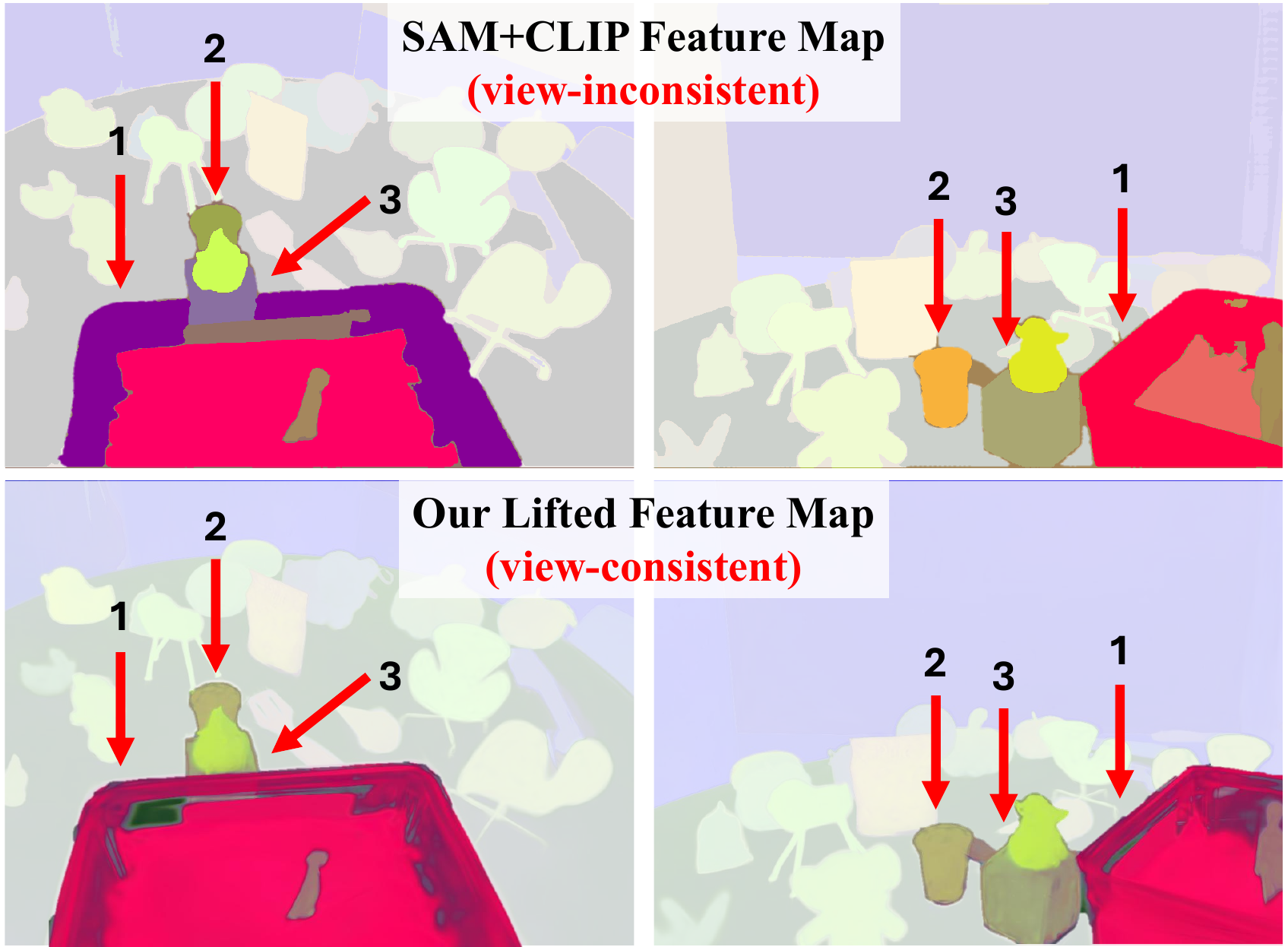}
    \caption{\textbf{Feature lifting.} The raw features from visual foundation models are not view-consistent. Feature lifting (\cref{sec:featurelifting}) alleviates this inconsistency.}
    \label{fig:main_multiview}
\end{figure}

%% file: figs/optimization.tex
\begin{figure*}[t!]
\centering
\includegraphics[width=0.995\textwidth]{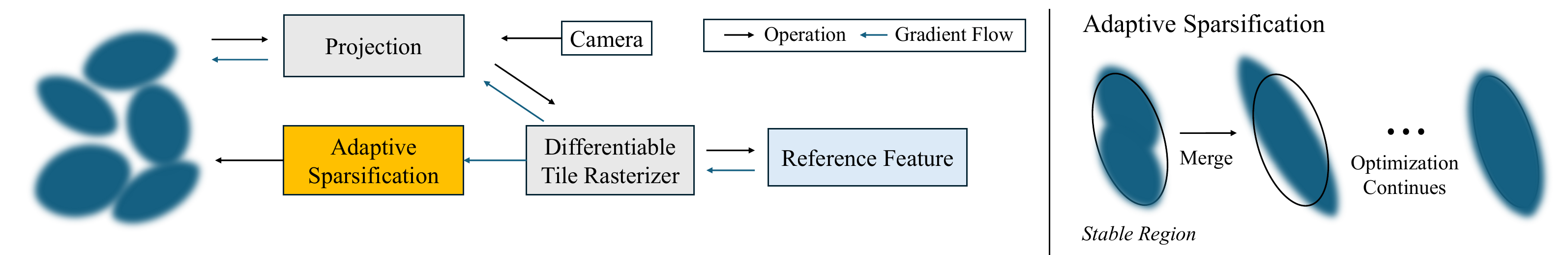}
    \vspace{-3mm}
    \caption{\textbf{Overview of our adaptive sparsification pipeline.} 
    Unlike the original 3D Gaussian Splatting, which preserves fine-grained details for photorealistic rendering, our method focuses on feature field reconstruction and merges redundant Gaussians to reduce unnecessary density, achieving effective sparsification.}
    \label{fig:optimization}
\end{figure*}

%% file: tables/algo.tex
\begin{algorithm}[t!]
\caption{Adaptive Sparsification} \label{alg:sparsification}
\begin{algorithmic}
\State $\boldsymbol{P},\boldsymbol{S},A \gets \text{Initial 3DGS position, scale, opacity}$
\State $\boldsymbol{F} \gets \text{lifted feature on 3DGS}$ \Comment{\cref{sec:featurelifting}}
\State Freeze feature lifted 3DGS $(\boldsymbol{P},\boldsymbol{S},\boldsymbol{F},A)$ 
\State $\boldsymbol{P}_{C},\boldsymbol{S}_{C},A_{C} \gets \boldsymbol{P},\boldsymbol{S},A$ \Comment{initialize \Ours{}}
\State $\textbf{Enc}, \textbf{Dec}$ \Comment{pretrained from lifted features $\boldsymbol{F}$}
\State $\boldsymbol{C}_{C} \gets \textbf{Enc}(\boldsymbol{F})$ \Comment{Encode features to color channels}
\State $i \gets 0$ \Comment{Iteration Count}
\While{$i < \text{MaxIteration}$}
\State $\boldsymbol{V} \gets \text{SampleTrainingView()}$
\State $\boldsymbol{I} \gets \text{Rasterize}(\boldsymbol{P}_{C}, \boldsymbol{S}_{C}, \boldsymbol{C}_{C}, A_{C}, \boldsymbol{V})$
\State $\boldsymbol{F}_{ref} \gets \text{RasterizeFeature}(\boldsymbol{P}, \boldsymbol{S}, \boldsymbol{F}, A, \boldsymbol{V})$
\State $\mathcal{L} \gets \text{Loss}(\textbf{Dec}(\boldsymbol{I}),\boldsymbol{F}_{ref})$
\State $(\boldsymbol{P}_{C}, \boldsymbol{S}_{C}, \boldsymbol{C}_{C}, A_{C}) \gets \text{Adam}(\nabla L)$
\If{IsPruneIteration($i$)}
\State \Call{PruneGaussians}{}
\EndIf
\If{IsMergeIteration($i$)}
\State \Call{MergeGaussians}{}
\EndIf
\State $i \gets i + 1$
\EndWhile
\end{algorithmic}

\vspace{1em}
\begin{algorithmic}
\Function{PruneGaussians}{}
\ForAll{Gaussians $g(\boldsymbol{\mu},\boldsymbol{\Sigma}, \boldsymbol{\alpha}, \boldsymbol{c})$}
\If{$C(g) < \tau_{con}$} \Comment{global contribution Eq.~\ref{eq:contrib}}
\State {PrunePoints($g$)}\Comment{prune gaussians}
\EndIf
\EndFor
\EndFunction
\end{algorithmic}

\vspace{1em}
\begin{algorithmic}
\Function{MergeGaussians}{}
    \ForAll{Gaussians $g_i(\boldsymbol{\mu}_i, \boldsymbol{\Sigma}_i, \alpha_i, \boldsymbol{c}_i)$}
        \If{$\nabla L < \tau_{grad}$} \Comment{gradient threshold}
            \ForAll{k-neighbors $g_j(\boldsymbol{\mu}_j, \boldsymbol{\Sigma}_j, \alpha_j, \boldsymbol{c}_j)$} \Comment{$j \ne i$}
                \State $\boldsymbol{d} \gets \boldsymbol{\mu}_j - \boldsymbol{\mu}_i$ \Comment{distance between Gaussians}
                \State $d_M \gets \max(\boldsymbol{d}^T \boldsymbol{\Sigma}_i^{-1} \boldsymbol{d},\ \boldsymbol{d}^T \boldsymbol{\Sigma}_j^{-1} \boldsymbol{d})$
                \If{$\langle \boldsymbol{c}_i, \boldsymbol{c}_j \rangle > \tau_{sim}$ \textbf{and} $d_M < \chi^2_\beta$}
                    \State \textbf{MergeGaussian}($g_i, g_j$) \Comment{~\Cref{eq:merge}}
                \EndIf
            \EndFor
        \EndIf
    \EndFor
\EndFunction

\end{algorithmic}
\end{algorithm}

%% file: sec/4_Experiments.tex
\section{Experiments}
\label{sec:experiments}
To evaluate our method, we conducted comparative experiments with other state-of-the-art feature-embedded 3D Gaussian splatting methods.
Further, we demonstrate the effectiveness of the feature-wise weighted averaging approach by applying it to both 3DGS~\cite{kerbl20233d} and LightGaussian~\cite{fan2023lightgaussian}.
We evaluate our method by measuring storage efficiency and performance on downstream tasks, including semantic segmentation and localization.
\input{figs/qual_lerf}

\subsection{Setup}
We use the widely adopted Replica~\cite{straub2019replica} and LERF~\cite{kerr2023lerf} datasets. 
We evaluate semantic segmentation on the Replica dataset using LSeg~\cite{li2022language} and MaskCLIP~\cite{zhou2022extract} across four scenes used by Feature-3DGS: room 0, room 1, office 3, and office 4.
Feature-3DGS~\cite{zhou2024feature} can embed the original feature directly into the 3D Gaussian splatting framework. It trains a computationally efficient $1\times 1$ decoder, a lower-dimensional feature can also be embedded into the 3D Gaussian splatting framework with minimal performance loss. 
We conducted experiments on Feature-3DGS with original, 128-dimensional, and 3-dimensional features to compare them with our compact and efficient representation. 
We then rendered the embedded features and computed similarity with text queries to obtain segmentation masks after thresholding. We measured the mean intersection-over-union (mIoU) and accuracy following the evaluation protocol~\cite{zhou2024feature}.
\input{tables/Replica_avg}

For the LERF~\cite{kerr2023lerf} dataset, we followed the LERF evaluation protocol and assessed mIoU and localization accuracy for four scenes: Ramen, Figurines, TeaTime, and Waldo Kitchen. 
We use the semantic features from LangSplat~\cite{qin2024langsplat} in this experiment. 
Since CLIP~\cite{radford2021learning} provides image-level features rather than pixel-level, LangSplat uses SAM~\cite{kirillov2023segment} to extract region-specific CLIP features.
These features are divided into whole, part, and subpart levels. Since our focus is on evaluating feature representations rather than the feature map granularity, we used the whole-level feature map consistently across all methods for a fair comparison.

On top of our method, we apply additional vector quantization following LightGaussian~\cite{fan2023lightgaussian} to compress the feature field even further.
We employed 3D Gaussian splatting scenes trained with 30k iterations. The same setup applies to Feature-3DGS and LangSplat in all experiments, including FPS measurements, conducted on a single NVIDIA RTX6000 Ada GPU.

\input{tables/Lerf_avg}
\subsection{Comparison}
\label{sec:comparison}

We conduct comparisons between \Ours{} and Feature-3DGS, using LSeg features on the Replica dataset. As shown in~\cref{tab:replica_avg}, our \Ours{} achieves competitive mIoU and accuracy while providing $121\times$ more compact 3D feature field than Feature-3DGS with a speed-up module. By employing adaptive sparsification to merge and prune unnecessary Gaussians, \Ours{} achieves comparable performance using fewer than $10$\% of the Gaussians. Additional vector quantization (\Ours{}+VQ) results in an even more compact 3D feature field, without notable performance degradation. In this experiment, we also incorporate the raw feature map as regularization.
\input{figs/qual_maskclip}
\input{tables/ablation_merge}

We then compare LangSplat, Feature-3DGS, and \Ours{} using the LERF dataset. We adopt the same feature map used in LangSplat. LangSplat compresses the 512-dimensional features to 3 dimensions using an autoencoder before lifting them to 3DGS, resulting in a more compact representation than Feature-3DGS.
In contrast, our per-Gaussian autoencoder, trained under the same feature distribution, leads to cleaner segmentation, as shown in~\cref{fig:qual_lerf}.
Consequently, as~\cref{tab:lerf_avg} indicates, our method achieves competitive performance while being more than $74 \times$ more compact than LangSplat and $245 \times$ more compact than Feature-3DGS. 
Particularly, when CLIP features are extracted for each segment using SAM masks, each region is represented by a single feature vector. In this case, our adaptive sparsification enables effective merging, allowing the 3D feature field to be described with only 5\% of the Gaussians compared to existing methods.
\input{tables/MaskCLIP}

The following experiment addresses a more general scenario than the previous two feature maps. 
LSeg, based on the DPT~\cite{ranftl2021vision} backbone, and CLIP with SAM both produce features at nearly the same resolution as the input image.
In contrast, MaskCLIP produces low-resolution, patch-level features, which lead to performance degradation in the baseline.
Our approach compensates for the limitations of these coarse features by using high-resolution reference features during adaptive sparsification.
As shown in~\cref{tab:maskclip} and~\cref{fig:qual_maskclip}, our method provides a representation over $182 \times$ more compact than Feature-3DGS, while achieving more than 30\% mIoU improvement and effectively removing noisy activations.

\subsection{Ablation}
We conducted an ablation study in~\cref{tab:combined_ablation} to demonstrate the effectiveness of each component of our pipeline. Ablations were performed for all experiments presented in~\cref{sec:comparison}. 
In particular, a key component of our method is the adaptive sparsification (\cref{sec:sparsification}) that eliminates redundant Gaussians. The merging step contributes to an additional 70\% storage reduction.
In addition, variance filtering (\cref{sec:featurelifting}) effectively removes noisy features from low-resolution features from MaskCLIP, contributing to improved performance.
After the feature compression stage, the number of Gaussians remains unchanged, but compressing high-dimensional features into a low-dimensional space contributes significantly to storage reduction.

\input{figs/3d_seg}
\input{tables/3DOVS_table}
\subsection{Open-vocabulary 3D Segmentation}
We additionally perform open-vocabulary 3D segmentation by directly querying the features embedded in the Gaussians. 
To associate \Ours{} with pre-trained 3DGS, we establish a mapping from \Ours{} to the pre-trained 3DGS after applying feature lifting (\cref{sec:featurelifting}). Each feature-lifted 3DGS point is mapped to its closest \Ours{} point by identifying the $k{=}3$ nearest neighbors in coordinate space and selecting the one with the highest cosine similarity in feature space. This allows us to propagate the text-based query results from \Ours{} back to the 3DGS for visualization.

We perform 3D segmentation on the 3D-OVS dataset ~\cite{liu2023weakly}. Specifically, the evaluation is conducted on the Office desk, Room, Snacks, and Sofa scenes included in the dataset. 
Unlike LangSplat~\cite{qin2024langsplat} and Feature-3DGS~\cite{zhou2024feature}, which train the autoencoder or decoder in 2D before lifting, our method learns the autoencoder directly on lifted 3D features, preserving the feature distribution between training and inference. As shown in~\cref{fig:cam_3dovs}, this leads to improved 3D segmentation performance. Open-vocabulary 2D segmentation results on the same dataset are also reported in \cref{tab:3dovs}.

To demonstrate the efficiency of our feature field representation, we conduct experiments on the large-scale outdoor KITTI-360 dataset~\cite{liao2022kitti}. Large-scale scenes pose a significant challenge for traditional optimization-based feature embedding due to their high computational cost. 
As shown in ~\cref{tab:kitti}, by leveraging a highly compact representation, \Ours{} substantially reduces storage overhead while enabling real-time rendering speeds. 
\cref{fig:cam_kitti} shows a visualization of the feature similarity between each Gaussian and a given text query. We compute the similarity directly between the embedded feature in each Gaussian and the text query feature, and map this similarity to a color for visualization. Importantly, this is based purely on the 3D Gaussian features, not on rendered features in 2D.
These results highlight the potential of \Ours{} for open-vocabulary semantic segmentation and localization in large-scale environments.
\input{figs/kitti}
\input{tables/kitti}

%% file: figs/qual_lerf.tex
\begin{figure*}[t!]
\centering
\includegraphics[width=0.95\textwidth]{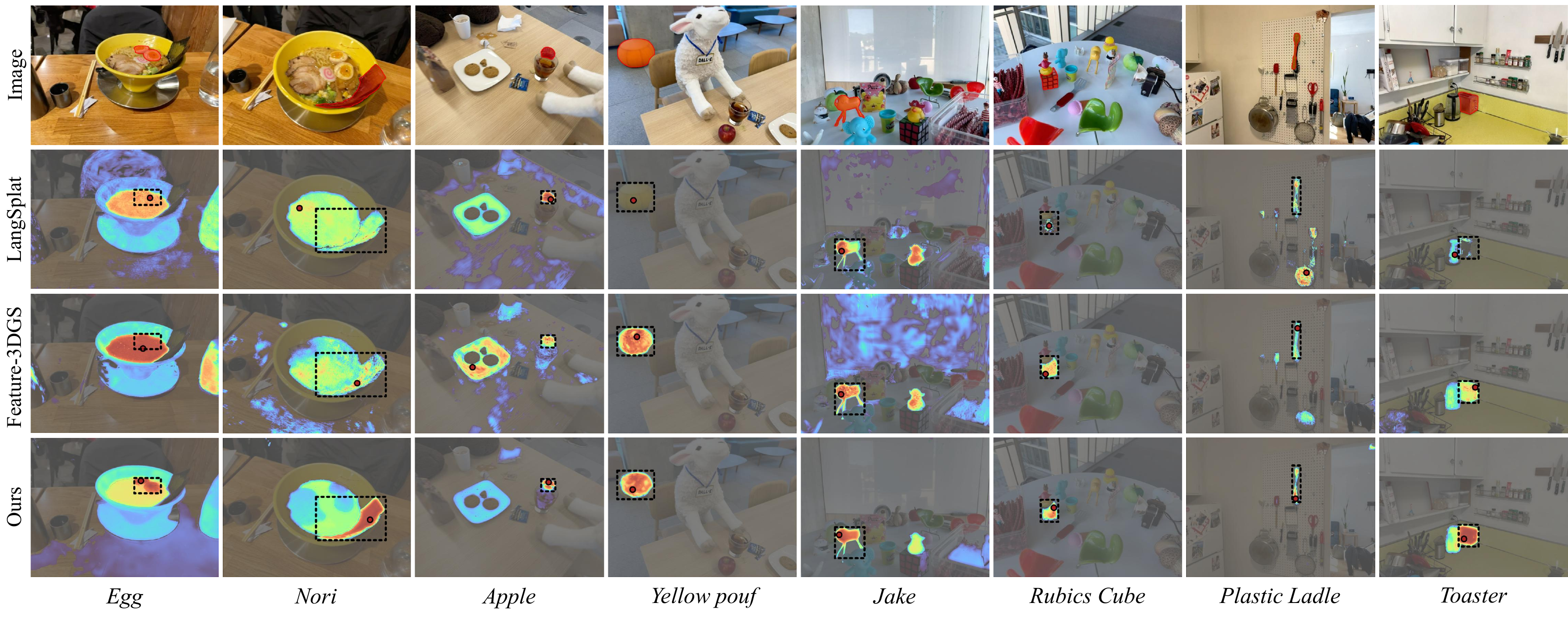}
    \vspace{-3mm}
    \caption{\textbf{Qualitative comparison.} We visualize open-vocabulary semantic segmentation and localization results using CLIP~\cite{radford2021learning} with SAM~\cite{kirillov2023segment} features on the LERF~\cite{kerr2023lerf} dataset. Our method shows precise results even for small objects in these tasks. Feature-3DGS~\cite{zhou2024feature} is tested with the speed-up module \textit{(128 dim)}. We overlay the ground truth segmentation for the query in red on the image for visualization.} 
    \label{fig:qual_lerf}
    \vspace{-3mm}
\end{figure*}

%% file: tables/Replica_avg.tex
\begin{table}[t!]
\centering
    \resizebox{0.93\columnwidth}{!}{%
    \setlength{\tabcolsep}{3pt}
    \begin{tabular}{l r r r r r}
    \toprule[1.2pt]
                                      & Storage$\downarrow$  & FPS$\uparrow$   & mIoU$\uparrow$  & Acc.$\uparrow$  & \#G$\downarrow$\\
    \midrule
    Feature-3DGS (512) & 1393.9M   & 7.2 & 73.0 & 91.9  & 636k  \\
    Feature-3DGS (128) & 463.9M & 113.8 & \textbf{73.4} & \textbf{92.9}   & 640k \\
    Feature-3DGS (3)   & 160.8M & 198.8 & 21.3 & 59.2   & 644k\\
    \midrule
    3DGS* & 1336.2M   & 6.8 & 70.1 & 90.9  & 600k  \\
    LightGaussian* & 458.8M   & 7.3 & 70.0 & 91.0  & 204k  \\
    \midrule
    \Ours{} (Ours) & 3.6M & \textbf{328.3}  & 70.8  & 91.6  & \textbf{47k}   \\
    \Ours{} + VQ (Ours) & \textbf{1.7M} & 327.3  & 70.1  & 90.9  & \textbf{47k}    \\
    \bottomrule[1.2pt]
    \end{tabular}
    }
    \vspace{-2mm}
    \caption{\textbf{Evaluation on Relica dataset with LSeg~\cite{li2022language}.} The asterisk (*) denotes results with feature lifting.}
    \label{tab:replica_avg}
 \vspace{-4mm}
\end{table}

%% file: tables/Lerf_avg.tex
\begin{table}[t!]
\centering
    \resizebox{0.95\columnwidth}{!}{%
    \setlength{\tabcolsep}{3pt}
    \begin{tabular}{l r r r r r}
    \toprule[1.2pt]
                                      & Storage$\downarrow$  & FPS$\uparrow$   & mIoU$\uparrow$  & Acc.$\uparrow$  & \#G$\downarrow$\\
    \midrule
    LangSplat & 314.9M   & 33.4 & 44.7 & 72.3  & 1270k  \\
    Feature-3DGS (128) & 1031.7M & 55.6 & 53.8 & 75.8   & 1423k \\
    Feature-3DGS (3)   & 345.6M & 90.6 & 4.3 &  3.5  & 1394k\\
    \midrule
    3DGS* & 2832.8M   & 1.7 & \textbf{56.7} & \textbf{85.4}  & 1289k  \\
    LightGaussian* & 986.0M   & 1.8 & 55.2 & 83.9  & 438k  \\
    \midrule
    \Ours{} (Ours) & 4.2M & \textbf{145.0}  & 52.4  & 76.8  & \textbf{55k}   \\
    \Ours{} + VQ (Ours) & \textbf{1.9M} & 144.3  & 51.7  & 75.7  & \textbf{55k}    \\
    \bottomrule[1.2pt]
    \end{tabular}
    }
    \vspace{-2mm}
    \caption{\textbf{Evaluation on LERF dataset with CLIP + SAM~\cite{radford2021learning,kirillov2023segment}.}}
    \label{tab:lerf_avg}
\end{table}

%% file: figs/qual_maskclip.tex
\begin{figure*}[t!]
\centering
\includegraphics[width=0.95\textwidth]{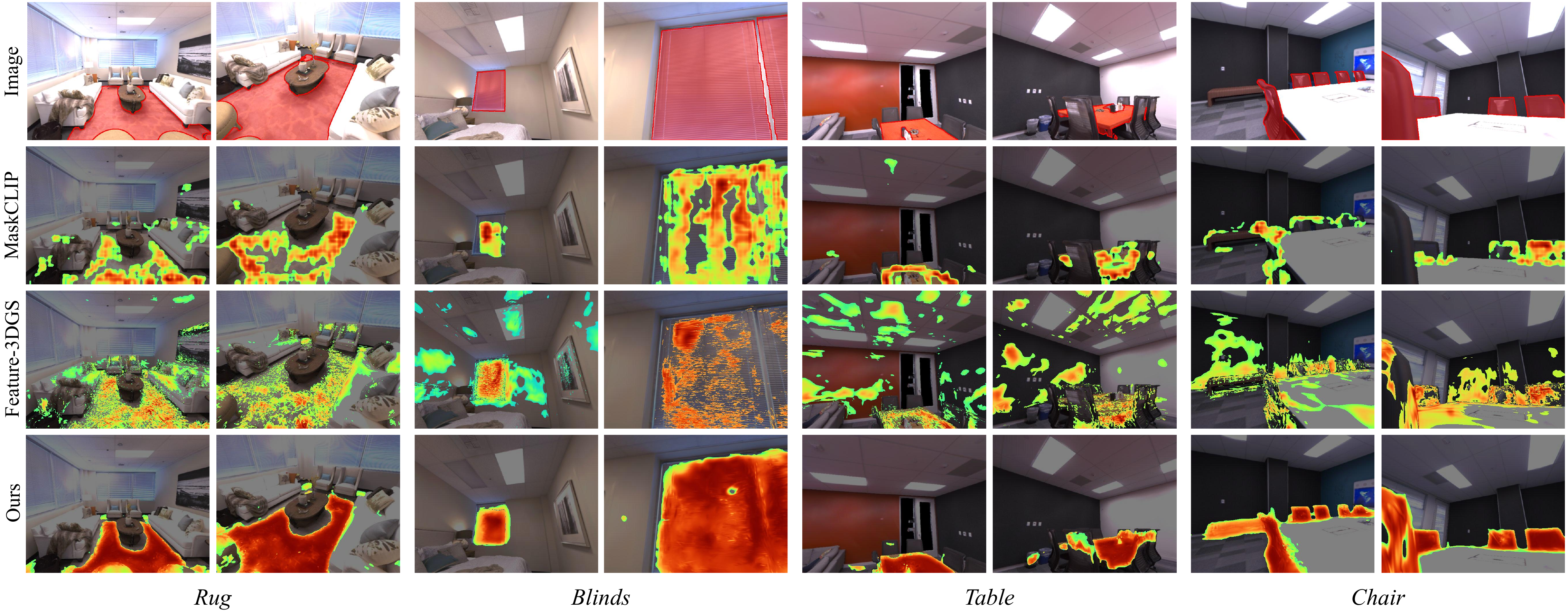}
    \vspace{-3mm}
    \caption{\textbf{Qualitative comparison.} We visualize open-vocabulary semantic segmentation results using MaskCLIP~\cite{zhou2022extract} features. Feature-3DGS~\cite{zhou2024feature} is tested with the speed-up module \textit{(128 dim)}. We highlight the ground truth masks in red for the corresponding query texts.}
    \label{fig:qual_maskclip}
    \vspace{-2mm}
\end{figure*}

%% file: tables/ablation_merge.tex
\begin{table*}[t!]
\centering
\scriptsize
\resizebox{0.9\textwidth}{!}{%
\setlength{\tabcolsep}{4pt}
\begin{tabular}{c c c | r r r r | r r r r r | r r r r r}
\toprule[1.2pt]
\multicolumn{3}{c|}{\textbf{Config}} &
\multicolumn{4}{c|}{\textbf{MaskCLIP (Replica)}} &
\multicolumn{5}{c|}{\textbf{LSeg (Replica)}} &
\multicolumn{5}{c}{\textbf{CLIP+SAM (LERF)}}  \\
VF & P & M &

Storage$\downarrow$ & FPS$\uparrow$ & mIoU$\uparrow$ & \#G$\downarrow$ &
Storage$\downarrow$ & FPS$\uparrow$ & mIoU$\uparrow$ & Acc$\uparrow$ & \#G$\downarrow$ &
Storage$\downarrow$ & FPS$\uparrow$ & mIoU$\uparrow$ & Acc$\uparrow$ & \#G$\downarrow$\\
\midrule
- & - & - &
42.5M & 245 & 42.8 & 600k &
42.6M & 254 & 61.0 & 87.6 & 600k &
90.7M & 101 & 29.7 & 57.8 & 1289k\\
\checkmark & \checkmark & - &
\cellcolor{colort}11.5M & \cellcolor{colort}311 & \cellcolor{colors}46.6 & \cellcolor{colort}152k &
\cellcolor{colort}12.1M & \cellcolor{colort}318 & \cellcolor{colorf}71.0 & \cellcolor{colors}92.0 & \cellcolor{colort}165k &
23.4M & 130 & \cellcolor{colors}53.4 & \cellcolor{colorf}77.9 & 324k \\
\checkmark & - & \checkmark &
27.6M & 279 & \cellcolor{colort}46.3 & 384k &
25.6M & 238 & \cellcolor{colors}70.9 & \cellcolor{colorf}92.1 & 355k &
\cellcolor{colort}20.4M & \cellcolor{colort}139 & \cellcolor{colort}52.4 & \cellcolor{colors}77.2 & \cellcolor{colort}284k \\
- & \checkmark & \checkmark &
\cellcolor{colors}3.0M & \cellcolor{colors}335 & 46.0 & \cellcolor{colors}36k &
\cellcolor{colorf}3.4M & \cellcolor{colors}324 & 69.8 & 91.5 & \cellcolor{colorf}43k &
\cellcolor{colors}4.2M & \cellcolor{colors}143 & \cellcolor{colorf}54.5 & 74.4 & \cellcolor{colors}56k \\
\checkmark & \checkmark & \checkmark &
\cellcolor{colorf}2.6M & \cellcolor{colorf}341 & \cellcolor{colorf}46.9 & \cellcolor{colorf}29k &
\cellcolor{colors}3.6M & \cellcolor{colorf}328 & \cellcolor{colort}70.8 & \cellcolor{colort}91.6 & \cellcolor{colors}47k &
\cellcolor{colorf}4.2M & \cellcolor{colorf}145 & 52.2 & \cellcolor{colort}76.8 & \cellcolor{colorf}55k \\
\bottomrule[1.2pt]
\end{tabular}%
}
\vspace{-2mm}
\caption{\textbf{Ablation study across all experiments.} Variance Filtering (VF), Pruning (P), Merging (M), \boxcolorf{First}, \boxcolors{Second}, \boxcolort{Third}}
\vspace{-2mm}
\label{tab:combined_ablation}
\end{table*}

%% file: tables/MaskCLIP.tex
\begin{table}[t!]
\centering
    \resizebox{0.95\columnwidth}{!}{%
    \setlength{\tabcolsep}{6pt}
    \begin{tabular}{l r r r r}
    \toprule[1.2pt]
    Metrics & Storage$\downarrow$ & FPS$\uparrow$ & mIoU$\uparrow$  & \#G$\downarrow$ \\
    \midrule
    MaskCLIP & - & - & 29.3  & -\\
    \midrule
    Feature-3DGS (512) &  1443.3M&  7.2 & 35.9 & 758k\\
    Feature-3DGS (128)& 474.8M & 118.3 & 33.7 & 760k\\
    Feature-3DGS (3)& 162.3M & 198.5 & 18.4 & 760k \\
     \midrule
    3DGS* & 1348.5M   & 7.2 & 46.3 & 600k  \\
    LightGaussian* & 448.3M & 7.4 & 46.2 & 204k \\
    \midrule
    \Ours{} (Ours)& 2.6M & 340.5 & 46.9 & \textbf{29k} \\
    \Ours{}+VQ (Ours)& \textbf{1.5M} & \textbf{342.3} & \textbf{47.1} & \textbf{29k} \\
    \bottomrule[1.2pt]
    \end{tabular}%
    }
    \vspace{-2mm}
    \caption{\textbf{Evaluation on Replica dataset with MaskCLIP~\cite{zhou2022extract}.}}
    \label{tab:maskclip}
 \vspace{-4mm}
\end{table}

%% file: figs/3d_seg.tex
\begin{figure}[t!]
\centering
\includegraphics[width=0.95\columnwidth]{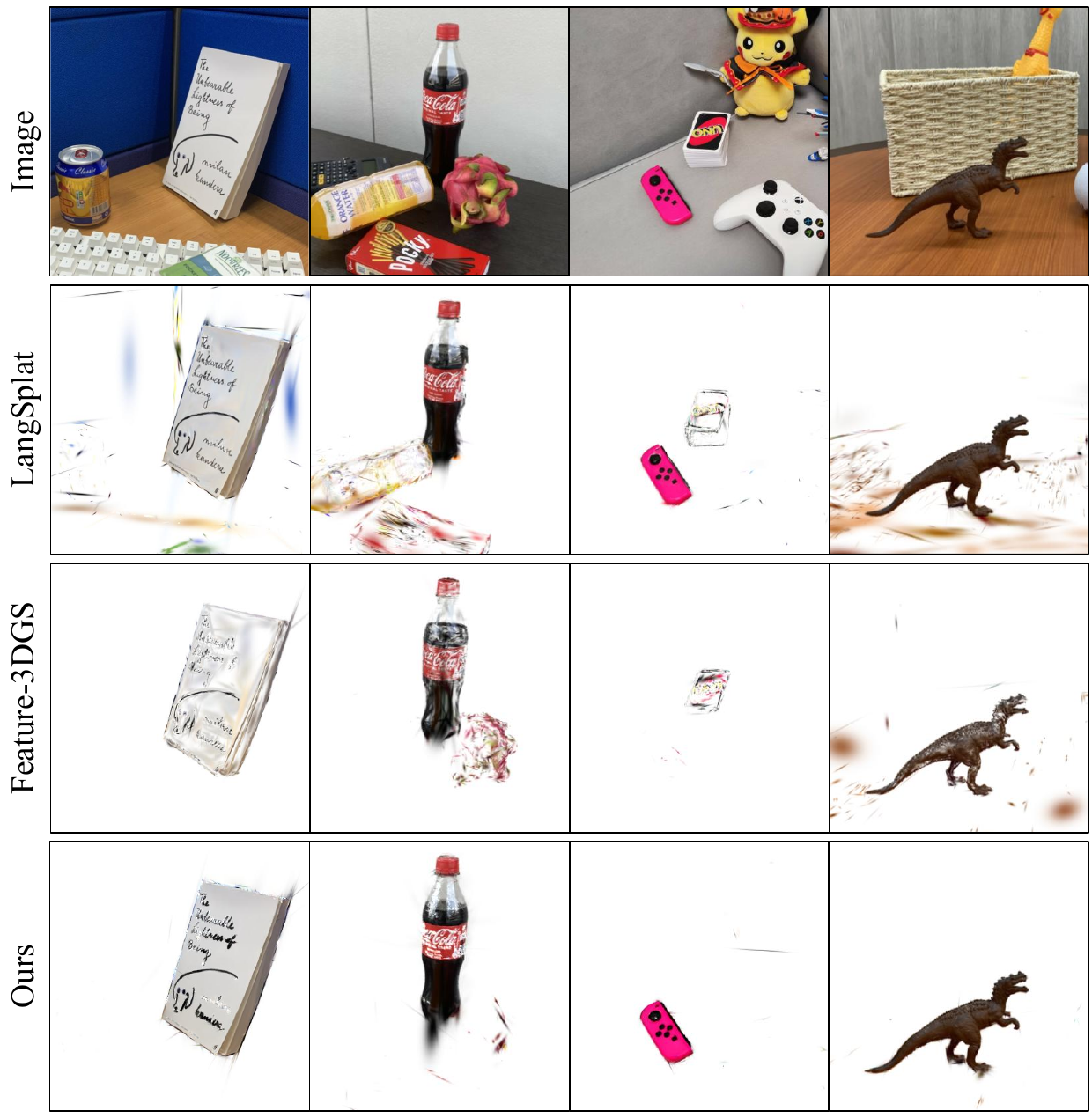}
    \caption{\textbf{3D Segmentation Results.} We perform open-vocabulary 3D segmentation on the 3D-OVS dataset. The following queries are used, in order: \textit{a book of The Unbearable Lightness of Being, Coca-Cola, a red Nintendo Switch Joy-Con controller, and Dinosaur.}}
    \label{fig:cam_3dovs}
\end{figure}

%% file: tables/3DOVS_table.tex
 \begin{table}[t!]
 \centering
     \resizebox{0.8\columnwidth}{!}{%
     \setlength{\tabcolsep}{3pt}
     \begin{tabular}{l r r r r}
     \toprule[1.2pt]
  & Storage$\downarrow$  & FPS$\uparrow$   & mIoU$\uparrow$ & \#G$\downarrow$\\
     \midrule
     Feature-3DGS(128) & 305.5M & 90.7 & 81.4 & 421k \\
     LangSplat  & 83.9M & 27.7 & 81.9 & 332k \\
     \midrule
     3DGS* & 746.8M & 2.6& 82.8 & 332k \\
     \midrule
     \Ours{} (Ours) & \textbf{1.7M} & \textbf{140.3} & \textbf{84.5} & \textbf{21k} \\
     \bottomrule[1.2pt]
     \end{tabular}
     }
     \vspace{-2mm}
     \caption{\textbf{Results on 3D-OVS Dataset with CLIP+SAM~\cite{radford2021learning,kirillov2023segment}.}}
     \label{tab:3dovs}
 \vspace{-4mm}
 \end{table}

%% file: figs/kitti.tex
\begin{figure}[t!]
\centering
\includegraphics[width=0.995\columnwidth]{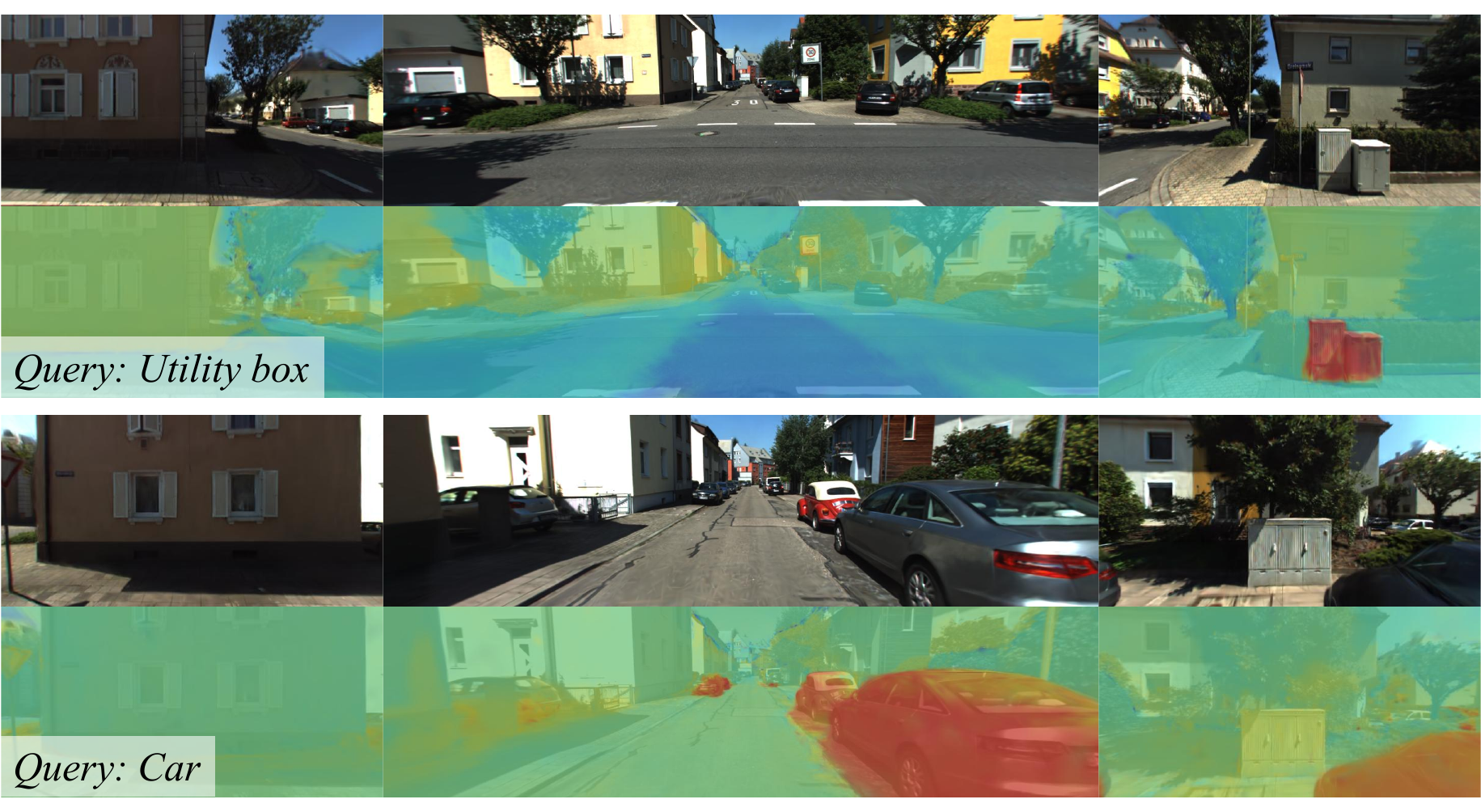}
    \caption{\textbf{Additional Result on KITTI-360 Dataset.} We visualize each Gaussian in \Ours{} based on its similarity to the text query and render the result. Blue indicates low similarity, while higher similarity is shown in red.}
    \label{fig:cam_kitti}
\vspace{-2mm}
\end{figure}

%% file: tables/kitti.tex
\begin{table}[t!]
\centering
\renewcommand{\arraystretch}{0.9} 
\setlength{\tabcolsep}{6pt} 
\resizebox{0.85\columnwidth}{!}{
\begin{tabular*}{\linewidth}{@{\extracolsep{\fill}}l r r r}
\toprule[1.2pt]
 & Storage$\downarrow$  & FPS$\uparrow$  & \#G$\downarrow$\\
\midrule
3DGS* & 3810.2M & 1.8  & 1734k \\
\Ours{} (Ours) & \textbf{6.2M} & \textbf{141.6} & \textbf{95k} \\
\bottomrule[1.2pt]
\end{tabular*}
}
\vspace{-2mm}
\caption{\textbf{Results on KITTI-360 Dataset.}}
\label{tab:kitti}
\vspace{-4mm}
\end{table}

%% file: sec/5_conclusion.tex
\section{Conclusion}
This paper presents a pipeline for constructing compact and fast 3D feature fields (\Ours{}). Unlike prior approaches, we train a per-Gaussian autoencoder on features lifted via weighted multi-view fusion. 
In addition, we propose an adaptive sparsification strategy that prunes and merges redundant Gaussians, reducing their count while maintaining representation fidelity.
Unlike other 3D feature field compression methods that store high-dimensional attributes separately and rely on auxiliary data structures such as hash grids, our method stores 3D features directly in the RGB channels of 3DGS, replacing color with features. This design makes it compatible with existing 3DGS pipelines.
While feature lifting is fast and efficient, the overall pipeline currently takes approximately 30 minutes per scene due to the autoencoder training and sparsification stages. We plan to accelerate these stages to minimize the overhead.

%% file: sec/X_supple.tex
\clearpage
\setcounter{page}{1}
\maketitlesupplementary
\appendix
\setcounter{table}{0}
\setcounter{figure}{0}
\renewcommand{\thefigure}{\Alph{figure}}
\renewcommand{\thetable}{\Alph{table}}

\begin{strip}
    \centering
    \includegraphics[width=0.95\textwidth]{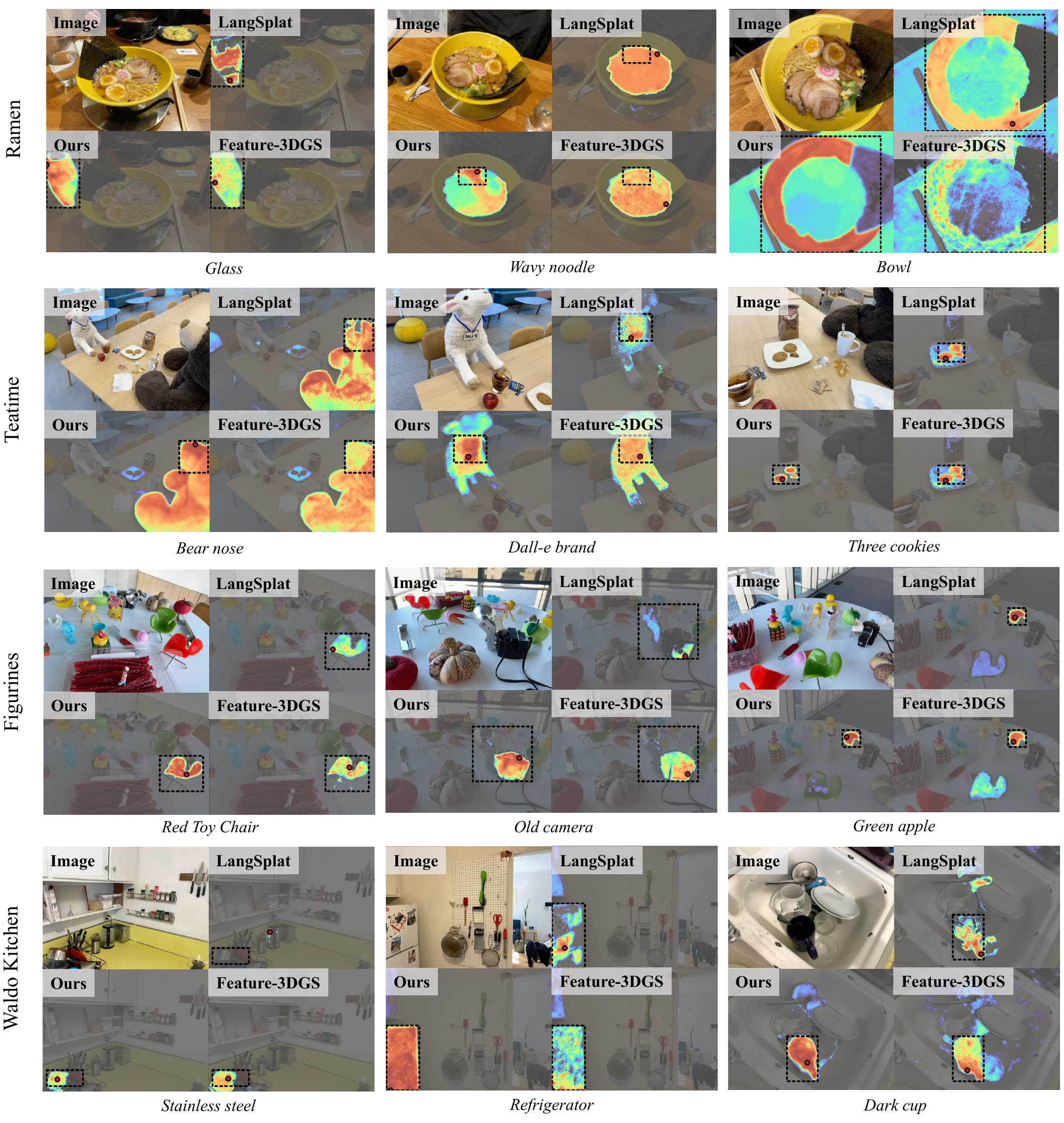}
    \captionof{figure}{\textbf{Additional Result on LERF Dataset.}}
    \label{fig:supp_lerf}
\end{strip}
\input{figs/supp_maskclip}
\input{figs/supp_kitti}
\clearpage
\input{sec/sup/implementation}
\section{Compatibility with 3DGS Compression}
While conventional 3DGS compression approaches focus on reducing storage for color attributes, our method targets feature representation and achieves higher compression efficiency.
For reference, \cref{tab:3dgs_zip} shows that \Ours{} achieves lower storage than efficient color 3DGS methods on the full MipNeRF360 dataset~\cite{barron2022mip}.
Therefore, our feature field can be combined with existing 3DGS compression methods~\cite{bagdasarian20243dgs,chen2025hac++,chen2024hac,navaneet2023compact3d,lee2025compression} to represent color and feature field jointly with little extra storage cost (for example, only 8.7 $+$ 2.5 $=$ 11.2MB is required when \Ours{}+VQ is stored with HAC++low).
\input{tables/3dgszip}

%% file: figs/supp_maskclip.tex
\begin{figure*}[t!]
\centering
\includegraphics[width=0.95\textwidth]{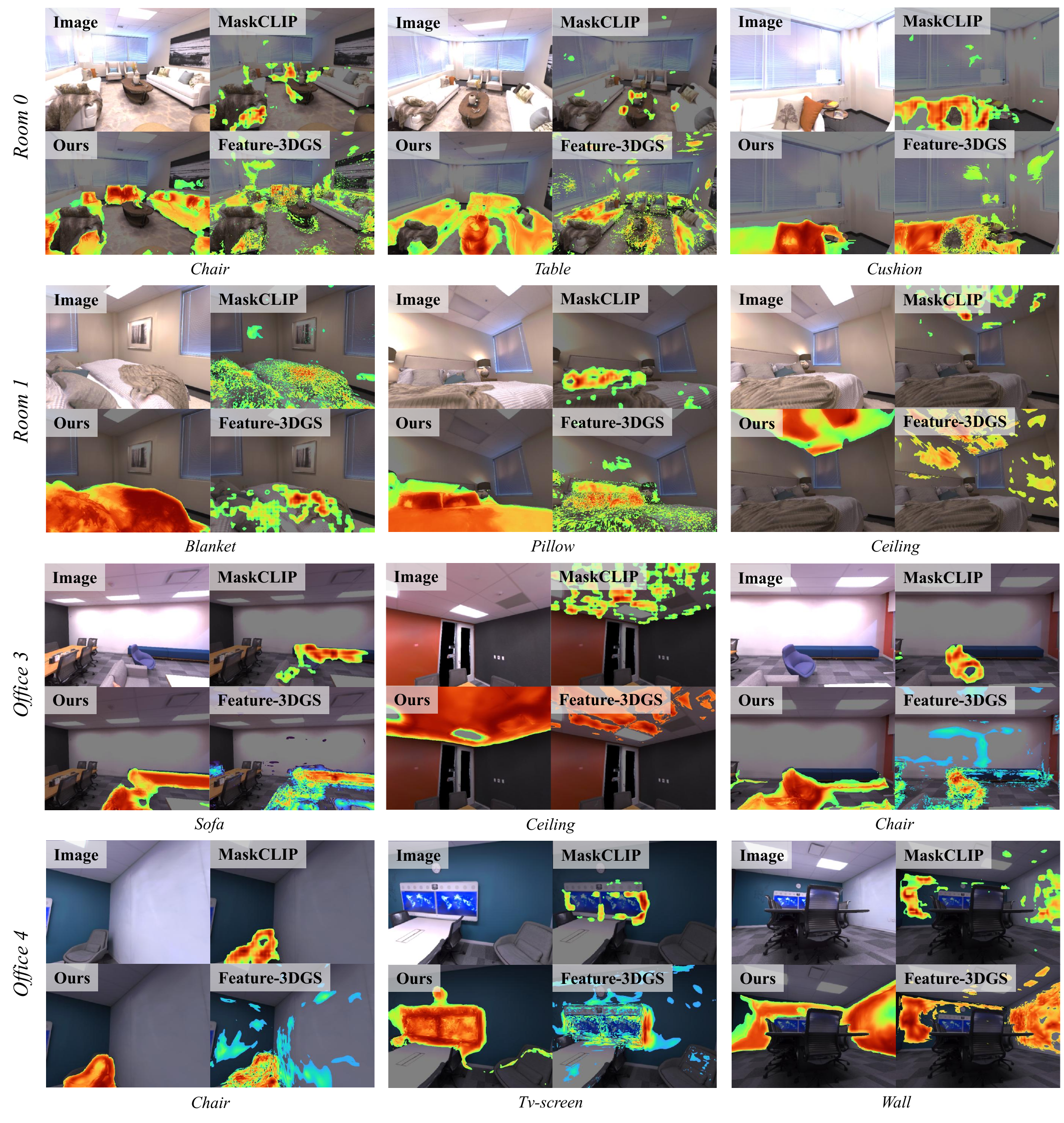}
    \caption{\textbf{Additional Result on Replica Dataset.}} 
    \label{fig:supp_lseg}
\end{figure*}

%% file: figs/supp_kitti.tex
\begin{figure*}[t!]
\centering
\includegraphics[width=0.95\textwidth]{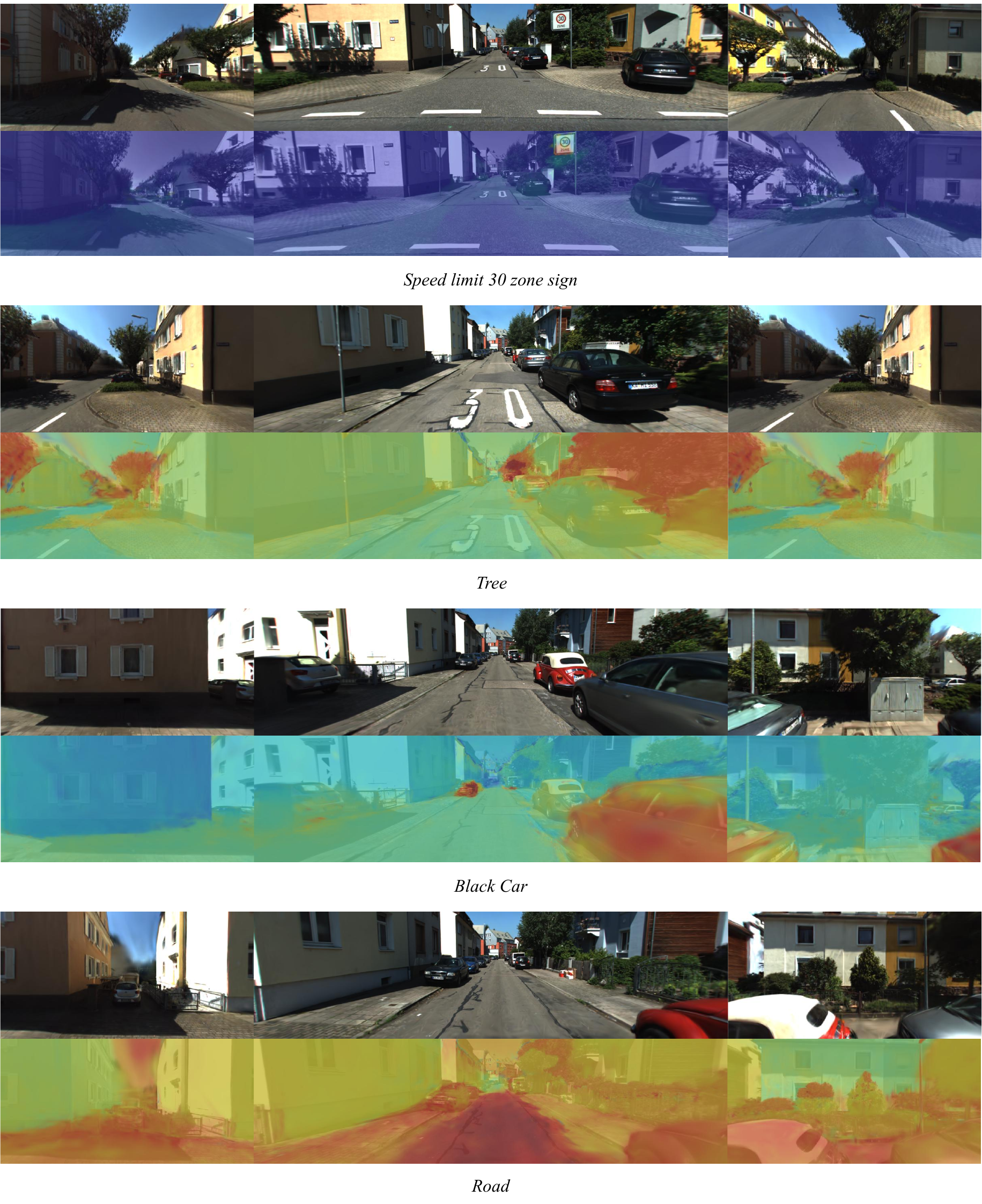}
    \caption{\textbf{Additional Result on KITTI-360 Dataset.}} 
    \label{fig:supp_kitti}
\end{figure*}

%% file: sec/sup/implementation.tex
\section{Additional Details}
\label{suppsec:implementation}

In MaskCLIP evaluation, we measured mIoU by selecting 5 to 6 categories among the labels provided with the replica gt segmentation map. 
The dataset used was Replica room\_0, room\_1, office\_3, and office\_4 for LSeg and MaskCLIP evaluation used by Feature-3DGS.
We used 3,000 iterations and a merge interval of 50. We set thresholds as $\tau_{con}=0.25$, $\tau_{sim}=0.999$, $\tau_{grad}=10^{-5}$, and $\chi_\beta^2=2.38$.


%% file: tables/3dgszip.tex
\begin{table}[h!]
\centering
\resizebox{\columnwidth}{!}{%
\setlength{\tabcolsep}{2pt}
\begin{tabular}{cccccc|c}
\toprule
 Compact3D & HAC-high & HAC-low & CodecGS & HAC++high & HAC++low  &{\centering \textbf{\Ours+VQ}}\\
\midrule
 18MB & 23MB & \cellcolor{colort}{16MB} & \cellcolor{colors}{10MB} & 19MB & \cellcolor{colorf}{8.7MB} &\textbf{2.5MB} \\
\bottomrule
\end{tabular}%
}
\caption{\textbf{Storage comparison with 3DGS.zip\cite{3DGSzip2025} results on MipNeRF360 dataset.} Baselines compress the 3DGS, which is designed for color representation.
In contrast, \Ours{} represents semantic features as a separate field, yet achieves smaller storage.}
\label{tab:3dgs_zip}
 \vspace{-2mm}
\end{table}